\documentclass[conference]{IEEEtran}
\IEEEoverridecommandlockouts
\usepackage{cite}
\usepackage{amsmath,amssymb,amsfonts}
\usepackage{algorithmic}
\usepackage{graphicx}
\usepackage{textcomp}
\usepackage{xcolor}
\usepackage{booktabs}
\usepackage{multirow}
\usepackage{verbatim}
\usepackage{array}
\usepackage{arydshln}
\def\BibTeX{{\rm B\kern-.05em{\sc i\kern-.025em b}\kern-.08em
    T\kern-.1667em\lower.7ex\hbox{E}\kern-.125emX}}
    
\usepackage{etoolbox}
\makeatletter
\patchcmd{\@makecaption}
  {\scshape}
  {}
  {}
  {}
\makeatother
\begin{document}

\title{An Unsupervised Feature Learning Approach to Reduce False Alarm Rate in ICUs\\
\thanks{This material is based upon work supported by the National Science Foundation under
Grant Number 1657260. Research
reported in this publication was also supported by the National Institute On Minority Health And Health
Disparities of the National Institutes of Health under Award Number U54MD012388.}
}

\author{Behzad Ghazanfari$^{1}$, Fatemeh Afghah$^{1}$, Kayvan Najarian$^{2,3,4}$, Sajad Mousavi$^{1}$, Jonathan Gryak$^{2}$, James Todd$^{1}$
\thanks{$^{1}$School of Informatics, Computing and Cyber Security (SICCS), Northern Arizona University, Flagstaff, AZ 86011}%
\thanks{$^{2}$Department of Computational Medicine and Bioinformatics, University of Michigan, Ann Arbor, MI, USA}%
\thanks{$^{3}$Michigan Center for Integrative Research in Critical Care, University of Michigan, Ann Arbor, MI, USA}%
\thanks{$^{4}$Department of Emergency Medicine, University of Michigan, Ann Arbor, MI, USA}%
}

\maketitle

\begin{abstract}
The high rate of false alarms in intensive care
units (ICUs) is one of the top challenges of using medical
technology in hospitals. These false alarms are often caused
by patients’ movements, detachment of monitoring sensors, or
different sources of noise and interference that impact the
collected signals from different monitoring devices. In this
paper, we propose a novel set of high-level features based on
unsupervised feature learning technique in order to effectively
capture the characteristics of different arrhythmia in electrocardiogram (ECG) signal and differentiate them from irregularity in signals due to different sources of signal disturbances. This unsupervised feature learning technique, first extracts a set of low-level features from all existing heart cycles of a patient, and then clusters these segments for each individual patient to provide a set of prominent high-level features. The objective of the clustering phase is to enable the classification method to differentiate between the high-level features extracted from normal and abnormal cycles (i.e., either due to arrhythmia or
different sources of distortions in signal) in order to put more
attention to the features extracted from abnormal portion of
the signal that contribute to the alarm. The performance of
this method is evaluated using the 2015 PhysioNet/Computing
in Cardiology Challenge dataset for reducing false arrhythmia
alarms in the ICUs. As confirmed by the experimental results,
the proposed method offers a considerable performance in terms
of accuracy, sensitivity and specificity of alarm detection only
using a few high-level features that are extracted from one single
lead ECG signal; while most of the reported works on false
alarm reduction on this public data set used several collected
signals including ECG lead II, ECG lead V, arterial blood
pressure (ABP) and photoplethysmogram (PPG) when they were
available.

\end{abstract}


\section{Introduction}
\label{sec:intro}

A key objective of the monitoring devices in ICUs is to constantly monitor patients' heart function to diagnose any life-threatening arrhythmia. 
An ECG signal measures the electrical activity of the heart and is known as an important tool in diagnosing different heart conditions, such as cardiac arrhythmia, ventricular hypertrophy, and myocardial infarction. 
In spite of many well-developed methods to detect abnormal rhythms, the ICUs still suffer from significantly high false alarm rates due to different reasons including complex nature of signal patterns for some arrhythmia, motion artifacts, noise, sensor detachment, and loose threshold settings of the monitoring devices \cite{lawless1994crying}. Such high false alarm rates negatively impact both patients and medical staff through desensitizing the medical staff to true alarms and increase the response time, rising to an issue commonly referred to as \textit{alarm fatigue}. Also, the frequent audio disturbance generated by false alarms can lead to sleep deprivation and depressed immune systems for the patients \cite{donchin2002hostile,imhoff2006alarm}.

Several research and clinical studies aimed at reducing the number of false alarms, ranging from expert systems that define several rules based on expert experiences to machine learning methods \cite{zhang2008patient,li2016signal,schmid2017reduction,he2015reducing}.  
One common drawback of such methods is their unstable performance on different datasets, meaning that they can show a promising performance on some datasets while presenting a poor performance on others, suggesting that the success of such methods highly depends on the characteristics of the data set used for training. While machine learning methods lead to better generalization and performance compared to expert system methods, there are several limitations in using typical machine learning methods in time-series data such as longitude ECG signals. A typical challenge in these methods is the chance of overfitting and inaccurate performance when dealing with a large number of noisy, redundant and correlated features extracted from time-series signals or their transform domains (e.g. wavelet). Feature selection / reduction methods attempt to reduce the large set of input features to the most salient ones, but they may result in discarding important features, thereby missing some meaningful patterns in the signals \cite{hira2015review,Afghah_NIPS,Afghah_Entropy}.

One distinction of the biomedical signals with other time-series datasets is the periodic /  semi-periodic behavior of such signals while the typical machine learning approaches are not capable of capturing this characteristic.  In this paper, we took advantage of this property and developed an unsupervised feature learning method that creates a set of low-dimensional features for each subject that captures important characteristics of the underlying patterns in high-dimensional time-series input data by putting an extra attention on the abnormal portions of the signal. In other words, this method constructs a set of higher level features that better captures the underlying patterns related to different alarm types by processing each patient's signal segment by segment.

Representation learning (feature learning) is one of most recent trends in machine learning that can improve the performance of machine learning methods by focusing on automatic discovery of features obtained from the raw data sets. Feature learning methods can be categorized into two groups of supervised and unsupervised learning. In supervised feature learning, the labels of input data are used to learn the feature representation step and train a method for classification \cite{oglic2016greedy}. In unsupervised feature learning, the input data without its labels is used to learn feature representation \cite{coates2011analysis,coates2012learning,goroshin2015unsupervised}. Unsupervised feature learning has been used in time-series data \cite{langkvist2014review,zhang2016unsupervised}; however, the current approaches have not provided the expected performance in ECG analysis yet, mostly due to presence of a wide set of diverse patterns related to different cardiac events. On the other hand, lack of availability of annotated long ECG recordings limits the application of supervised feature learning methods.

To the best of our knowledge, the previously reported unsupervised feature learning methods for ECG analysis are based on deep learning \cite{eduardo2017ecg}. In \cite{zhang2017patient}, clustering is used to find the most representative beats for training recurrent neural networks. In \cite{xu2018towards}, the authors used aligned
heartbeats for deep neural network to learn features and classification.

In this paper, an unsupervised feature learning method is proposed that takes the segmented unannotated ECG signals as input, clusters these segments to learn the relationships among these segments, and then uses the resulting clusters toward learning high-level features. Specific to the detection of false alarms, the algorithm first segments the ECG signal of each patient based on its heartbeats. These segments contain important features that represent the characteristics of the major ECG components such as P wave, QRS complex, and T wave. Then, the extracted segments of each patient's ECG signal are clustered into several clusters. 
Each cluster represents a group of segments that are the most similar to each other. In fact, we build a bidirectional top-down and down-top feature learning by multi-resolution features that helps to better focus on patterns from toppest to lowest ones and vice versa (lowest to toppest). In the lowest to toppest direction, it provides a framework to capture non-linear relations, differences and similarities among those local features in the higher resolutions (i.e., segments and clusters). 

It is worth noting that in this method we only utilize one-lead ECG signal collected from the patients and achieved comparable results to the methods that utilized all collected signals available in 2015 PhysioNet challenge data set (i.e. ECG lead II, ECG lead V, arterial blood pressure (ABP) or photoplethysmogram (PPG). This fact suggests the potential capability of this method to be used for cardiac event detection in  wearable heart monitoring systems, since the majority of these remote monitoring systems such as Holter monitor only collect one lead ECG signal.

The rest of paper is structured as follows: In Section \ref{date_description}, the database used in this study is introduced. 
Section \ref{methodology} presents the proposed method. In Section \ref{experiment}, the experiment results are described, followed by the concluding remarks in Section \ref{conclusion}.

\section{Database Description}
\label{date_description}

In this paper, a publicly available database for `Reducing False Arrhythmia Alarms in the ICU' provided by PhysioNet/Computing in Cardiology Challenge 2015 \cite{PhysioNet,clifford2015physionet} is used. Each recording includes one or two ECG leads and one or more pulsatile waveforms, such as ABP and  PLETH. This dataset focuses on five life-threatening arrhythmia alarms of Asystole (ASY), Extreme Bradycardia (EBR), Extreme Tachycardia (ETC), Ventricular Tachycardia (VTA), and Ventricular Flutter/Fibrillation (VFB). The training set contains 750 recordings, and the test dataset is not available to public; therefore, we used the training dataset for both training and testing purposes by utilizing k-fold cross validation. This study focuses only on the ECG lead II signal as this lead is the only recording available for the majority of patients. The ECG signals are 5 minutes in length and have been re-sampled at the rate of 250 Hz.  Out of the 750 subjects, 29 of them were removed since they did not include ECG lead II signal. Table \ref{tab:StatDataset} shows the statistics of the numbers of true and false alarms for each arrhythmia type considered in this study. 

\begin{table}[h!]
  \begin{center}
    \caption{Number of true and false alarms for each arrhythmia type.}
     \vspace{-10pt}
    \label{tab:StatDataset}
    \begin{tabular}{l|c|c|c} 
      \textbf{Alarm} & \textbf{\# of Patients} & \textbf{False Alarm} & \textbf{True Alarm}\\
       \hline
      ASY & 116 & 94 & 22\\
      EBR & 86 & 41 & 45\\
      VFB & 57 & 51 & 6\\
      ETC & 131 & 8 & 123\\
      VTA & 331 & 245 & 86\\
      \hline
      Total & 721 & 439 & 282\\
    \end{tabular}
  \end{center}
  \vspace{-20pt}
\end{table}

\section{Methodology} \label{methodology}
The majority of methods for signal processing that are based on time-series analysis or transform-based techniques handle the entire collected signal in the same way, while empirically the abnormal portions of the signals often contain more information about the event of interest. A common feature of several biomedical signals is the existence of a basic periodic  pattern that can help distinguish between a normal- and an abnormal condition of a physiological system. For instance, in ECG signal analysis, the periodic normal ECG signal reveals several basic information about the heart function such as the heart beat, while the abnormal ECG can help in diagnosis of several arrhythmias. The multi-resolution feature extraction method presented below leverages this difference between the normal and abnormal portions by extracting additional higher resolution features from the abnormal sections, thereby enhancing the accuracy of arrhythmia detection process. 

\begin{figure}[htb]
\centering
  \includegraphics[height=0.2\textheight,keepaspectratio]{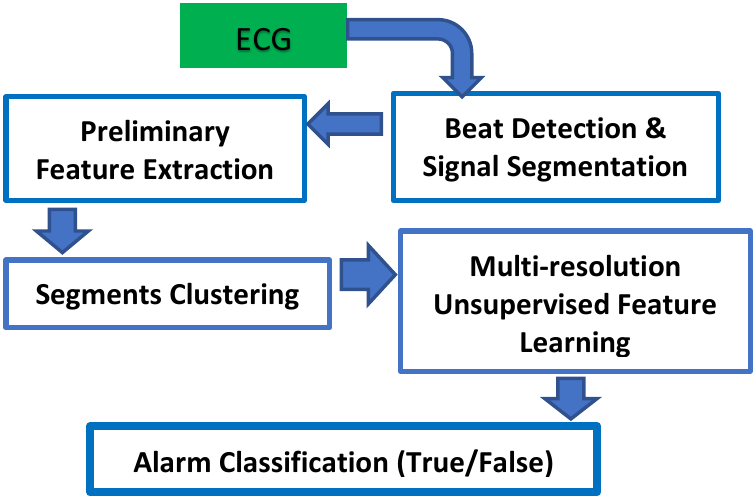}
  \caption{A schematic diagram of the proposed method.}
  \label{fig:flowchart} 
\end{figure}

Figure \ref{fig:flowchart} provides a schematic diagram of the overall method. First, the ECG signals are segmented by performing peak detection and decomposing each signal into P-T segments. Then, a set of preliminary features are extracted from each segment as described in Table \ref{tab:ECG_related_feats}. After that, the detected segments for each patient are clustered into several clusters using the $K$-means algorithm, and a set of multi-resolution features are learned from each cluster in an unsupervised way. $K$-means algorithm is used for clustering as it provides a fast and robust performance in most applications. Finally, these new constructed features are normalized to classify the alarms as true or false.

\subsection{Beat Detection and Signal Segmentation}

\begin{figure}[htp]
\centering
  \includegraphics[width=\linewidth,height=0.8\textheight,keepaspectratio]{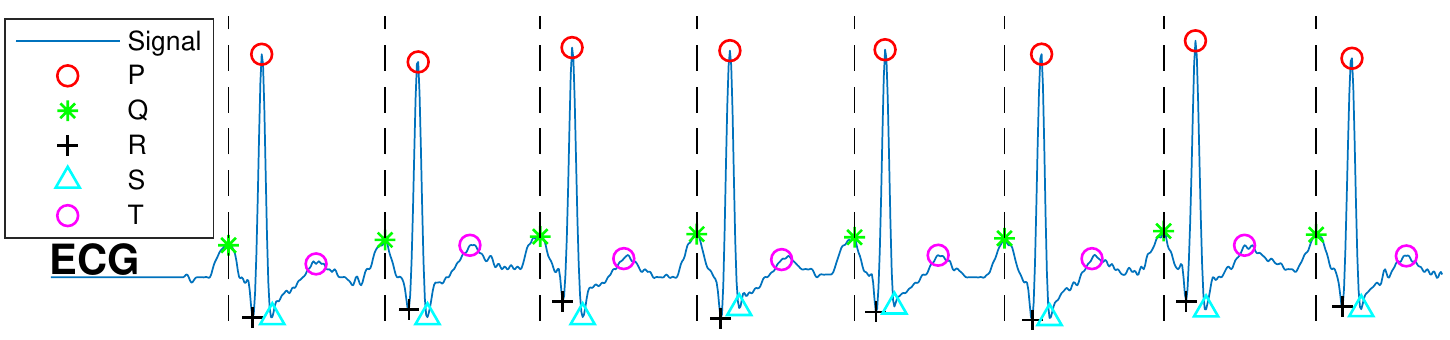}
  \caption{The detected peaks and computed segments of ECG signal. Each part between each two hashed-lines is considered as one segment.} 
  \label{fig:ecg_a}
\end{figure}

There are several techniques to segment the ECG signals and extract the beat-to-beat intervals. Pan Tompkins algorithm is one of the most common,  and low-computational methods to detect the QRS complex in the ECG signals \cite{pan1985real}. Most of the proposed methods for ECG segmentation are highly sensitive to several disturbances such as noise, interference, and motion artifacts. The objective of the proposed method is to decrease the number of false alarms due to these abnormalities in the signal. Let us consider three general cases of possible segments in the ECG signal including the normal cycles, arrhythmia segments and the abnormal ones (i.e., noisy segments, the irregular segments due to inaccurate segmentation, or the ones affected by sudden changes in QRS amplitudes). The proposed feature learning technique can significantly degrade the impact of inaccuracy in segmentation, noise, and abrupt changes in alarm detection, since these distortions lead to different behavior in segments compared to the known arrhythmia patterns. During the clustering phase, different abnormal segments due to the aforementioned sources of distortions are more likely to be clustered into separate clusters. Then, the classification technique can recognize such abnormal ones by learning the relations between the labels and the extracted clusters. We should note that the proposed feature learning algorithm is a generic method and can be applied over any segmentation techniques such as Hamilton-Tompkins and Hilbert transform-based algorithm. After detecting R-peaks, the presence of other waves in the signal including P, Q, S and T are detected using adaptive searching windows for each peak. Then, each segment is identified from the onset of its P wave to the offset of its T wave. Figure \ref{fig:ecg_a} illustrates an ECG signal annotated with R-peaks, P-, QRS-, and T-waves.

\subsection{Preliminary Feature Extraction}
As mentioned earlier, the entire 5-minute-long recordings of ECG lead II are utilized during feature extraction. Preliminary features are extracted from each segment of the given ECG signal in the time domain. These preliminary features can be categorized into three groups, 1) $x$- and $y$-coordinates of each present waves (i.e., P, Q, R, S, and T waves) of the signal, for example $P_{x}$ and $P_{y}$; 2) the intervals between the beginnings of different waves, for example, RR intervals; and 3) the intervals of the amplitudes between the waves or RR amplitude intervals. The set of extracted ECG-related features (84 in total) are provided in Table \ref{tab:ECG_related_feats}. \textcolor{black}{The $x$ values in each segment are defined as the relative distance to the location of $Q$-wave as a reference point}. The $y$-value refers to the amplitude of the ECG signal. In the first category of primary features, the $(x,y)$ of P,Q,R,S, and T are extracted for each segment. For example, $P_x$ and $P_y$ refers to $x$ and $y$ coordinate of the P wave. In the second category of primary features, notation  $OnQRS\_x$ refers to the average of $x$ values of P and Q waves and the average of $x$ values of S and T is defined as $OFFQRS\_x$. Notations $OnQRS\_y$ and $OnQRS\_y$ are the corresponding values of ECG signal for $OnQRS\_x$ and $OFFQRS\_x$, respectively. In the third category of primary features, the interval (difference between the x values of two waves) are measured. For example, $PP interval$ means the difference between the $x$ value of the P of the current segment and the $x$ value of the P of the next segment. $RR2$ interval means the distance between the $x$ value of R of the current segment and $x$ value of R-peak of the next-but-one segment. In the fourth category of primary features, the difference of the $y$-values of two peaks are measured. For example, $R-R amplitude$ means the difference between the $y$ value of the R-peak of the current segment and the $y$ value of the R-peak of the next segment. $R-R2 amplitude$ means the difference between the $y$ value of R-peak of the current segment and the R-peak of the next-but-one segment.

\begin{table}[htb]
  \begin{center}
    \caption{The preliminary 84 features extracted for each segment}
    \label{tab:ECG_related_feats}
    \begin{tabular}{c|c|c|l} 
      \textbf{No.} & \textbf{Feature} &  \textbf{No.} & \textbf{Feature} \\
       \hline
      1 & Px &  11 & OnQRS\_x \\ 
      2 & Py &  12 & OnQRS\_y \\ 
      3 & Qx &  13 & OFFQRS\_x \\
      4 & Qy &  14 & OFFQRS\_y \\
      5 & Rx &  $\cdots$ &  $\cdots$ \\
      6 & Ry &  $\cdots$ &  $\cdots$ \\
	  7 & Sx &  80 & RR2 interval \\
      8 & Sy &  82 & RR interval \\
      9 & Tx &  83 & R-R2 amplitude\\
      10 & $\cdots$& 84 & R-R amplitude \\
    \end{tabular}
  \end{center}
\end{table}

\subsection{Multi-resolution Unsupervised Feature Learning}
The proposed method learns the patterns of different heart arrhythmias through several high-level features, which are learned thorough unsupervised feature learning. The key contribution of the proposed method is to put attention on abnormal portions of the longitude ECG signals, either arrhythmia or distorted portions, through clustering.  When the clustering approach  is applied on the segments, the center of each cluster is a representative of the segments that belong to that cluster. Then, the proposed method assigns different weights to the formed clusters based on the number of segments within each clusters. These weights represent the probability of a cluster occurrence. The high-level learned features are built based on the position of the centers of these clusters as well as the number of their encountered segments.

The extracted segments of the signals are the units that can provide micro resolution frame of the signal to distinguish the encountered abnormalities in different clusters. Let us denote a cluster $t$ with $c_{t}$ as a set of segments as $\{ Seg_{1}, Seg_{2}, ..., Seg_{n}\}$, with $|c_{t}|=n$.
The segments' clusters provide a macro (higher-level) representation of the signals in which the center of each cluster represents the behavior of a set of segments that are most similar to each other. Therefore, the center of a cluster, centroid, is used as the representative of encountered segments. The number of features of the center of clusters is the same as the number of features of each segment (i.e., 84). The center of cluster $t$ is denoted with $cc_{t}$  in which $cc_{t}=\langle \text{f}_{t1}, \text{f}_{t2},..., \text{f}_{t84} \rangle$. The number of clusters' elements, the center of clusters, and the distance among these centers are defined as high-level  features that capture the relations among different signals. For instance, when the number of elements of a cluster is relatively small and its centers is far from other clusters for a patient, and the clusters similar to this cluster are frequently seen in other patients signals, this cluster likely represents an arrhythmia. Therefore, these high-level features are used  to construct a low-dimensional set from the high-dimensional low-level features of segments.

The preliminary features of centers are normalized between 0 and 1 for each cluster. Also, the clusters of each patient are ordered based on the number of their members in an ascending order. The high-level features can be categorized in four categories as follows:
\begin{itemize}
\item The patient's heart rate and alarm type (i. e., \textit{Tachycardia}, \textit{Asystole}, \textit{Ventricular Flutter Fib}, \textit{Ventricular Tachycardia}, and \textit{Bradycardia}). 

\item The number of elements of each cluster ($|c_{t}|$) in an ascending order.  For example, for the case of having two clusters with $|c_{1}|< |c_{2}|$, there are two features in the following order order $|c_{1}|$ and $|c_{2}|$.

\item The ratio of summation of features' values for the centers of each cluster to the number of elements in that cluster based on the order $ \dfrac{\sum_{i=1}^{84}cc_{ti}}{|c_{t}|}$. For example, if there are two clusters, we will have two values in the following order of $ \dfrac{\sum_{i=1}^{84}cc_{1i}}{|c_{1}|}$ and $ \dfrac{\sum_{i=1}^{84}cc_{2i}}{|c_{2}|}$, assuming that $|c_{1}|< |c_{2}|$.

\item The ratio of summation of features' values for the centers of each cluster to the number of all segments of that patient based on the order, $ \dfrac{\sum_{i=1}^{84}cc_{ti}}{\sum_{t=1}^{k}|c_{t}|}$. For instance, for the case of two clusters, we will have two values in the following order of $ \dfrac{\sum_{i=1}^{84}cc_{1i}}{|c_{1}|+|c_{2}|}$ and $ \dfrac{\sum_{i=1}^{84}cc_{2i}}{|c_{1}|+|c_{2}|}$, assuming that $|c_{1}|< |c_{2}|$.
\end{itemize}

\subsection{Model Building}
The learned high-level features of each signal are used in two different scenarios as the input for the classification phase. Here, we used two well-known classification methods of \textit{Boosted Trees} and \textit{RUSBoosted Trees} from the \textit{MATLAB's Classification Learner APP} since they are based on ensemble learning and  provide robust results \cite{matlab2018learningapp}. However, the proposed method is independent of the choice of the classification technique and the extracted high-level features can be fed to any classifiers. 

In the first scenario, only the unsupervised learned features are used as the input. In the second scenario, the unsupervised learned features are added to the features that are extracted using discrete wavelet transform, as described in Section \ref{experiment}. We also evaluated the proposed method using two different distance measures. As shown in Section \ref{experiment}, we achieved a great performance in alarm classification by only using one signal of ECG lead II and even by using a few number of high-level features that indicates the capability of this method in arrhythmia detection.

\section{Experiment Results}
\label{experiment}

\begin{table*}[t]
\caption{Comparison of the performance of Boosted Trees classification method for different scenarios of using i) the learned high level features using our proposed method when using \textit{cityblock} distance metric in clustering (31 features) referred as $HLF_{\text{cityblock}}$, ii) the learned high level features using our proposed method when using \textit{squared Euclidean} distance metric in clustering (31 features) referred as $HLF_{\text{Euclidean}}$ , iii) $120$ statistical and information-theoretic features extracted using 6-level DWT, and iv) 588 low-level features extracted from the last 10 second of the signal, referred as $LLF$. }
\label{table:results1}
\vskip 0.15in
\begin{center}
\begin{small}
\begin{sc}
\begin{tabular}{l||c||c||cc||cc}
\toprule
 & \multicolumn{6}{c}{\textit{Boosted Trees}}\\

\hdashline

{Scenarios} & \textit{$LLF$} &  \textit{DWT} & \textit{$HLF_{\text{cityblock}}$} & \textit{$HLF_{\text{Euclidean}}$} & \textit{DWT + $HLF_{\text{cityblock}}$}  & \textit{DWT + $HLF_{\text{Euclidean}}$}  \\\hdashline
\midrule
 Accuracy  & 0.791 &  0.717 & \textbf{0.818}  & 0.793 & 0.788 & 0.803 \\\hdashline
 Specificity   & 0.79 & 0.65 & \textbf{0.83}  & 0.79 & 0.77 & 0.78 \\\hdashline
 Sensitivity     & 0.80 & 0.76 & 0.81  & 0.80 & 0.80 & \textbf{0.82}\\\hdashline
 AUC   & 0.82 & 0.78 & \textbf{0.85}  & 0.82 & 0.84  & \textbf{0.85}\\\hdashline
\bottomrule
\end{tabular}
\end{sc}
\end{small}
\end{center}
\vskip -0.1in
\end{table*}

\begin{table*}[t]
\caption{Comparison of the performance of RUSBoosted Trees classification method for different scenarios of using i) the learned high level features using our proposed method when using \textit{cityblock} distance metric in clustering (31 features) referred as $HLF_{\text{cityblock}}$, ii) the learned high level features using our proposed method when using \textit{squared Euclidean} distance metric in clustering (31 features) referred as $HLF_{\text{Euclidean}}$ , iii) $120$ statistical and information-theoretic features extracted using 6-level DWT, and iv) 588 low-level features extracted from the last 10 second of the signal, referred as $LLF$. }
\label{table:results2}
\vskip 0.15in
\begin{center}
\begin{small}
\begin{sc}
\begin{tabular}{l||c||c||cc||cc}
\toprule
 & \multicolumn{6}{c}{\textit{RUSBoosted Trees}}\\

\hdashline

{Scenarios} & \textit{$LLF$} &  \textit{DWT} & \textit{$HLF_{\text{cityblock}}$} & \textit{$HLF_{\text{Euclidean}}$} & \textit{DWT + $HLF_{\text{cityblock}}$}  & \textit{DWT + $HLF_{\text{Euclidean}}$}  \\\hdashline
\midrule
 Accuracy    & 0.775 & 0.725 & \textbf{0.779} & 0.77     & 0.753   & 0.771 \\\hdashline
 Specificity & 0.73 &  0.63  & \textbf{0.75}  & 0.72& 0.67 & 0.70 \\\hdashline
 Sensitivity & 0.80 & 0.80   & 0.80  & 0.80         & 0.81 & \textbf{0.82} \\\hdashline
 AUC         & 0.83 & 0.79   & 0.82      & 0.81         & 0.83 & \textbf{0.85}  \\\hdashline
\bottomrule
\end{tabular}
\end{sc}
\end{small}
\end{center}
\vskip -0.1in
\end{table*}

The performance of this method is evaluated using the 2015 PhysioNet computing in cardiology challenge dataset \cite{PhysioNet,clifford2015physionet}.
We should note that in this study we only used the ECG lead II to extract the features as this signal was the most common available recordings in the database. However, the majority of existing studies on this dataset used all available signals for the patients including ECG II, ECG V, PPG, and ABP (for many patients only two of these signals are available) and yet our proposed method achieves a comparable performance related to these works. We would also like to note that the winner projects of 2015 Physionet challenge reported the results based on using the public portion of the data set for training and the private part for test, however, in this work, the public training set is used for both test and training with K-fold cross-validation.

To evaluate the performance of our proposed method, several evaluation metrics such as accuracy, sensitivity,  specificity, and area under the ROC curve (AUC) of the proposed method are reported for the two aforementioned scenarios. The results of this method is compared against two common methods of time-domain and wavelet-domain analysis of the ECG signal \cite{DWT_False}. In the wavelet-based method, the discrete wavelet transform (DWT) is applied on the entire 5-minute ECG lead II recordings, where a 6-level Daubechies 8 (db8) wavelet is used since there is a good match between its shape with the shape of the ECG signal.
Because feeding all wavelet coefficients as features into the classification algorithm can result in over--fitting; we reduce the number of features by extracting 20 representative statistical and information--theoretic features from each level of the wavelet vectors as summarized in Table 2 in \cite{Afghah_Entropy}. In this case, the overall number of wavelet-based features is $20 \times{6}=120$. We also compared the performance of the proposed method against a scenario, where the low-level $84$ features per segments are extracted for the last $10$ seconds of the ECG recording, as the data set identifies that the cardiac events have occurred  within the 10 seconds of the signal. Since these 10-second recordings include a different number of segments for each patient, we considered seven last segments of the signal as an average number of observed segments for the last 10-seconds and extracted the $84$-features per segment for these last segments (total of  $84\times{7}=588$ low-level features). We should note that one can consider a longer interval of the signal in time domain which results in a larger number of low-level features and can result in over-fitting.  

Tables \ref{table:results1} and \ref{table:results2} present the accuracy, sensitivity, specificity and the AUC of two classification techniques of Boosted Tree and RUSBoosted Trees, respectively, for five scenarios including, i) using the $588$ low-level features extracted from the last 10 seconds of the signal, ii) using the $120$ statistical and information-theoretic features extracted from a 6-level DWT over the entire 5 minutes ECG lead II signal, iii) only using the proposed high level features when the distance metric of $K$-means clustering is \textit{cityblock}, iv) only using the proposed high level features when the distance metric of $K$-means clustering is \textit{Squared Euclidean distance}, and v) both the proposed high-level learned features as well as the DWT-based features. The number of clusters in the $K$-means clustering is considered as five. As the results show, \textit{Boosted Trees} working with only the proposed features with distance metric of \textit{cityblock} provides better performance in nearly all measures in comparison to using only the DWT-based features as well as using both sets of features high-level features and the DWT-based ones together (due to potential redundancy between the two sets), and also compared to the case of using \textit{squared Euclidean distance} metric. 

In summary, the proposed method obtains a better results compared with the common wavelet-based and time-domain based methods by only  $31$ high-level features since the unsupervised learned features have better discrimination powers while they need less much computation resources than commonly used methods.  

\section{Conclusion}
\label{conclusion}

In this paper, an unsupervised feature learning approach is proposed for analysis of periodic/semi-periodic biomedical signals. This method works based on a clustering approach and learns a few number of high-level features from  several formed clusters of time-domain segments of the signals. The learned high-level features can handle the various patterns and variations in ECG signals in a autonomous and scalable way by splitting the signals to their constituent segments and representing the discrimination of the segments by clustering. In this paper, the proposed feature learning technique has been applied on a single lead ECG signal available in 2015 PhysioNet challenge to distinguish the patterns in ECG signal associated to several arrhythmia's from other potential distortions in ECG signals due to noise, interference, motion artifacts and other source of disturbance in collected ECG signals. For this purpose, we applied the proposed method on the entire 5-minutes available ECG recordings to enable the model to learn both normal and abnormal patterns in the collected signal per patient. As seen in the experimental results, the proposed method is capable of achieving a better performance compared to common false alarm detection based on DWT and a time-domain analysis- using the last 10 seconds of the signal that include the event- by only using a few number of high-level features and a low level of computations.   

\bibliographystyle{IEEEtran} 
\bibliography{reference}

\end{document}